# Designing Kernel Scheme for Classifiers Fusion

Mehdi Salkhordeh Haghighi
Computer Department
Ferdowsi University of Mashhad, Iran
haghighi@ieee.org

Abedin Vahedian
Computer Department
Ferdowsi University of Mashhad, Iran
vahedian@um.ac.ir

Hadi Sadoghi Yazdi
Computer Department
Ferdowsi University of Mashhad, Iran
sadoghi@sttu.ac.ir

Hamed Modaghegh
Electrical Eng Department
Ferdowsi University of Mashhad, Iran
modaghegh@yahoo.com

*Abstract* — In this paper, we propose a special fusion method for combining ensembles of base classifiers utilizing new neural networks in order to improve overall efficiency of classification. While ensembles are designed such that each classifier is trained independently while the decision fusion is performed as a final procedure, in this method, we would be interested in making the fusion process more adaptive and efficient.

This new combiner, called Neural Network Kernel Least Mean Square[1], attempts to fuse outputs of the ensembles of classifiers. The proposed Neural Network has some special properties such as Kernel abilities, Least Mean Square features, easy learning over variants of patterns and traditional neuron capabilities. Neural Network Kernel Least Mean Square is a special neuron which is trained with Kernel Least Mean Square properties. This new neuron is used as a classifiers combiner to fuse outputs of base neural network classifiers. Performance of this method is analyzed and compared with other fusion methods. The analysis represents higher performance of our new method as opposed to others.

## I. INTRODUCTION

Classification is the process of assigning unknown input patterns of data to some known classes based on their properties. For a long time, many research areas in designing classifiers have focused on improving efficiency, accuracy and reliability of classifiers for a wide range of applications. Fusing outputs of base classifiers as an ensemble working in parallel on input feature space is an attractive method to build more reliable classifiers. It is well known that in many situations, combining outputs of several classifiers leads to improved classification results. This occurs because each classifier produces error on a different area of the input space. In other words, the subset of input space that each classifier labels correctly will differ from one classifier to another. This implies that by using information from more than one classifier, it is probable that a better overall accuracy is obtained for a given problem. On the other hand, Instead of picking up just one classifier, a better approach would be to use more than one classifier while averaging their outputs. The new classifier might not be better than the single best classifier but will diminish or eliminate the risk of picking up an inadequate single classifier.

Combining classifiers is an established research area based on both statistical pattern recognition and machine learning. It is known as committee of learners, mixtures of experts, classifier ensembles, multiple classifier systems, consensus theory, etc. By having a number of different classifiers; it is wise to use them in a combination in the hope of increasing the overall accuracy and efficiency. It is intuitively accepted that classifiers to be combined should be diverse. If they were identical, no improvements would result in combining them. Therefore, diversity among the team has been recognized as a key point. Since the main reason for combining classifiers is to improve their performance, there is clearly no advantage to be gained from an ensemble that is composed of a set of identical classifiers or classifiers that show the same patterns of generalizations.

In principle, a set of classifiers can vary in terms of their weights, the time they take to converge, and even their architecture, yet constitute the same solution and present the same patterns of error when they are tested [1]. Obviously, when designing an ensemble, the aim is to find classifiers which generalize differently (different diversity) [2]. There are a number of parameters which can be manipulated with this goal in mind: initial conditions of the architecture, training data, and training algorithm of the base classifiers.

The reminder of this paper is organized as follows. A brief review of related works in Section 2 is followed by the structure of our new classifier combiner explained in Section 3. Results of employing the classifier combiner on some known benchmarks in along with comparing these results to those obtained by other known classifiers are presented in Section 4. Concluding remarks and conditions under which our new classifier combiner is expected to perform well are discussed in Section 5.

## II. BACKGROUND

There are two main categories in combining classifiers: fusion and selection. In classifier fusion, each ensemble member has knowledge of the whole feature space. In classifier selection, on the other hand, each ensemble member knows well a part of the feature space and is responsible for objects in this part. Therefore, different types of combiners are used for each method. Moreover, there are combination schemes lying between the two principle strategies. Combination of fusion and selection is also called competitive/cooperative classifier or ensemble/modular approach or multiple/hybrid topology [3, 4].

---

[1] NNKLMS

*A. Classifiers fusion taxonomy*

In fusion category, the possible ways of combining outputs of classifiers in an ensemble depend on the information obtained from the individual members [5]. A general categorization based on types of outputs of classifiers in the ensemble is fused according to labeled outputs together with fusion of continuous valued outputs. A number of methods based on labeled outputs are Majority vote, weighted majority vote, Naive Bayes Combination, Behavior Knowledge Space Method, Wernecke's Method and SVD[2], as shown in Figure 1.

The methods introduced for fusion of continuous valued outputs are divided in two general categories: Class Conscious Combiners and Class Indifferent Combiners. Some combiners do not need to be trained after the classifiers in the ensemble have been trained individually while other combiners need additional training. These two types of combiners are called trainable/non-trainable combiners or data dependent/data independent ensembles. Some of Class indifferent combiners are Decision Templates and Dempster Shafer. Some other evolutionary methods are also proposed in the literature in the category of trainable combiners.

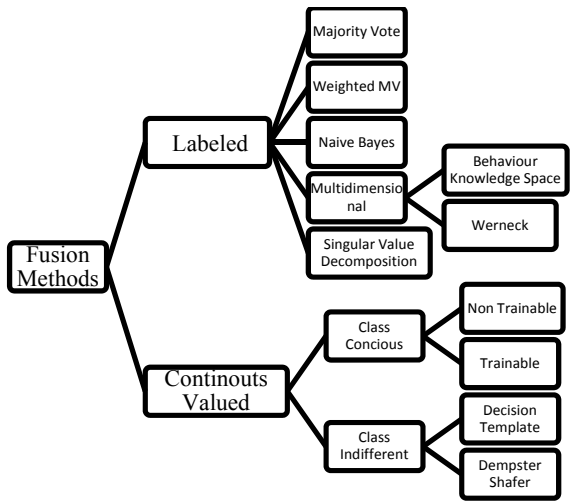

Figure 1: Classifiers fusion methods

The main idea in classifier selection is an "oracle" that can identify the best expert for a particular input x. This expert's decision is accepted as the estimated decision of the ensemble for x. Two general approaches are proposed in selection: Decision-Independent Estimate or priory approach and Decision-Dependent Estimate or posteriori approach. In Decision-Independent Estimate the competence is determined based only on the location of x, prior to finding out what labels are suggested for x by the classifiers. In Decision-Dependent Estimates the class predictions for x by all the classifiers are known. Some of the algorithms proposed on these approaches are direct k-NN Estimate, Distance-Based k-NN Estimate and Potential Functions Estimate.

From another point of view, there are two other methods of gaining optimized combiners. One method is based on choosing and optimizing the combiner for a fixed ensemble of base classifiers called decision optimization or non generative ensembles. The other method creates ensemble of diverse base classifiers by assuming a fixed combiner called coverage optimization or generative ensembles. Combination of these methodologies is also used in applications as decision/coverage optimization or non generative/generative ensembles [6, 7].

*B. Classifiers fusion methods*

Once an ensemble of classifiers has been created, an effective way of combining their outputs should be found. Amongst the methods proposed the majority vote is by far the most simple and popular approach. Other voting schemes include the minimum, maximum, median, average, and product [8, 9]. The weighted average approach evaluates optimal weights for the individual classifiers and combines them accordingly [11]. The BKS[3] method selects the best classifier in some region of the input space, and bases its decision on the best classifier's output. Other approaches include rank-based methods such as the Borda count, Bayes approach, Dempster–Shafer theory, decision template [10], fuzzy integral, fuzzy connectives, fuzzy templates, probabilistic schemes, and combination by neural networks [12].

The combiner could also be viewed as a scheme to assign data independent or data dependent weights to classifiers [13, 14, and 15]. The boosting algorithm of Freund and Schapire [16] maintains a weight for each sample in the training set that reflects its importance. Adjusting the weights causes the learner to focus on different examples leading to different classifiers. After training the last classifier, the decisions of all classifiers are aggregated by weighted voting. The weight of each classifier is a function of its accuracy.

---

[2] Singular Value Decomposition

[3] Behavior Knowledge Space

A layered architecture named stacked generalization framework, has been proposed by Wolpert [17]. The classifiers at Level 0 receive as input the original data, and each classifier outputs a prediction for its own sub problem. Each layer receives as input the predictions immediately preceding layer. A single classifier at the top level outputs the final prediction. Stacked generalization is an attempt to minimize the generalization error by making classifiers at higher layers to learn the types of errors made by the classifiers immediately below them.

Another method of designing classifiers combiner based on fuzzy integral and MSVM[4] was proposed by Kexin Jia [18]. The method employs multi-class support vector machines classifiers and fuzzy integral to improve recognition reliability. In another approach, Hakan, et al [19] proposed a dynamic method to combine classifiers that have expertise in different regions of input space. The approach uses local classifier accuracy estimate to weight classifier outputs. The problem is formulated as a convex quadratic optimization problem, which returns optimal nonnegative classifier weights with respect to the chosen objective function, and the weights ensure that locally most accurate classifiers are weighted more heavily for labeling the query sample.

*C. Training strategies*

The difficulties arising in combining a set of classifiers are evident if one considers the metaphor of a committee of experts. While voting might be the way such a committee makes final decision, it ignores their differences in skills and seems pointless if the constitution of the committee is not carefully set up. This may be solved by assigning areas of expertise and following the best expert for each new item of discussion. In addition, the experts may be asked to provide some confidence. But, the claim of an expert to have a great insight with respect to a problems not shared by anyone else, could dominate the decision at points that seem arbitrary for the others. This makes the decision of fake or expert fairly hard. The problem, therefore cannot be detected if the established committee is given a decision procedure to use their own confidence and follow their decisions. An optimal decision procedure would, therefore, require evaluating the committee which means supplying problems with known solutions, studying expert's advice and constructing the combined decision rules. In terms of classifiers this is called training [20] which is needed unless the collection of experts fulfills certain conditions. Instead of using one of the fixed combining rules, a training set can be used to adapt the combining classifier to the classification problem. A few possibilities will be discussed.

There are a number of methods for training classifier combiners. The Stacking method [21] was one of the first learning methods for classifier combiners. In this method a meta-level classifier is trained using the outputs of the base-level classifiers using the probabilities of each of the class values returned by each of the base level classifiers [22]. Another stacking approach based on meta decision trees have also been proposed [23]. Some of the typical approaches for building classifier combiners are Bagging [24, 25], Random subspace [26], Rotation forest [27] and different values for each classifier parameters [28].

Moreover, there are some evolutionary methods for building classifier combiners [29]. A method introduced by Loris Nanni and Alessandra Lumini [30] uses a genetic-based version of the correspondence analysis for combining classifiers. The correspondence analysis is based on the orthonormal representation of the labels assigned to the patterns by a pool of classifiers. Instead of the orthonormal representation, they used a pool of representations obtained by a genetic algorithm. Each single representation is used to train different classifiers; these classifiers are combined by vote rule.

The performance of the classifier combiner relies heavily on the availability of a representative set of training examples. In many practical applications, acquisition of a representative training data is expensive and time consuming. Consequently, it is not uncommon for such data to become available in small batches over a period of time. In such settings, it is necessary to update an existing classifier in an incremental fashion to accommodate new data without compromising classification performance on old data. One of the incremental algorithms proposed by Robi Polikar [31] named Learn++, is an algorithm for incremental training of NN pattern classifiers which enables supervised NN paradigms, such as MLP[5], to accommodate new data, including examples that correspond to previously unseen classes. Furthermore, the algorithm requires no access to previously used data during subsequent incremental learning sessions, yet at the same time, it does not forget previously acquired knowledge. Learn++ utilizes ensemble of classifiers by generating multiple hypotheses using training data sampled according to carefully tailored distributions. The outputs of the resulting classifiers are combined using a weighted majority voting procedure. Robi Polikar et al revised the Learn++ algorithm in 2009 and introduced Learn++.NC [32].

---

[4] Multi-Class Support Vector Machines

[5] Multilayer Perceptron

*D. NN classifier as combiner*

In the taxonomy of classifier types, Neural Networks have been used as a classifier for a long time due to efficiency, flexibility and adaptability. In fusion applications, they play important role as classifiers combiner. Neural classifiers can be divided into relative density and discriminative models depending on whether they aim to model the manifolds of each class or to discriminate the patterns of different classes [52]. Examples of relative density models include auto-association networks [53,54] and mixture linear models [51, 52 and 55]. Relative density models are closely related to statistical density models, and both can be viewed as generative models. Discriminative neural classifiers include the MLP, RBF[6] net, the PC[7] [56, 57], etc. The most influential effort in artificial neural networks learning algorithm is the development of BP8 algorithm which has two major shortcomings– the training may be getting stuck in local minima and the convergence could be slow.

LMS[9] is one of these learning methods. It is an intelligent simplification of the gradient decent method for learning [58] using the local estimate of the mean square error. In other words, the LMS algorithm is supposed to employ a stochastic gradient instead of the deterministic gradient used in the method of steepest decent. While LMS algorithm can learn linear pattern very well, it does not extend to nonlinear. To overcome this problem Puskal [60] used kernel method [59] and derived an LMS algorithm directly in kernel feature space and employed the kernel trick to obtain the solution in the input space. The Kernel LMS algorithm provides a computational simple and an effective algorithm to train nonlinear systems. Pulskal also showed that KLMS[10] have good result in non-linear time series prediction and non-linear channel modeling and equalization [60].

## III.  KLMS BASED COMBINER

We first introduce a novel neuron with logistic activation function. Classification is done in high dimensional kernel feature space. The neuron exploits kernel trick like KLMS to train itself. This classifier is non-parametric and therefore can discriminate every nonlinear pattern without predefined parameters. Therefore, structure of the neuron is analyzed before analyzing the structure of the new combiner system.

*A. LMS Algorithm*

In 1959 the LMS algorithm was introduced as a simple way of training a linear adaptive system with mean square error minimization. An unknown system $Y_{(n)}$ is to be identified and the LMS algorithm attempts to adapt the filter $\widehat{Y_{(n)}}$ to make it as close as possible to $Y_{(n)}$. The algorithm uses $u_{(n)}$ as input, $d_{(n)}$ as desired output and $e_{(n)}$ as calculated error. LMS uses steepest-descent algorithm to update the weight vector so that the weight vector converges to optimum Wiener solution. Updating weight vector based on equation (1) is applied:

$$w_{(n+1)} = w_{(n)} + 2\mu \times e_{(n)} \times u_{(n)} \qquad (1)$$

Where w(n) is weight vector and μ is step size and u(n) is input vector. The filter output Y is calculated by equation (2):

$$Y_{(n)} = \overline{w} \times u_{(n)} \qquad (2)$$

Successive adjusting of the weight vector eventually leads to the minimum value of the mean squared error.

*B. Kernel tricks Methods*

Kernel methods are applied to map input data into a HDS[11]. In HDS various methods can be used to find linear relations between input data. Mapping procedure is handled by Φ functions shown in Figure 2. By kernel methods[12], it is possible to map data to a high dimensional feature space known as Hilbert space. At the heart of KM is a kernel function that enables KM to operate at the mapped feature space without even computing coordinates at that high dimensional space. This is done with the aid of famed kernel trick. Any kernel function must satisfy mercer conditions. Many algorithms are developed that work based on KM such as KLMS [33, 34], based on LMS algorithm.

Kernel methods are also applied successfully in classification, regression problems and more generally in machine learning (SVM[13] [35], regularization networks [36], K-PCA[14] [37], K-ICA [15][38]). Kernel methods have been used to extend linear adaptive filters expressed in inner products to nonlinear algorithms [41, 39, and 40]. Pokharel et al. [41, 42] applied this "kernel trick" to the least mean square[16] algorithm [43, 44] to obtain a nonlinear adaptive filter in RKHS[17], which have joined KLMS. Kernel functions help the algorithm to handle the converted input data in the HDS even without knowing coordinates of the data in that space. This is done simply by computing the kernel of input data instead of calculating the inner products between images of all pairs of data in HDS. This method is called the kernel trick [36].

---

[6] Radial Basis Function
[7] Polynomial Classifier
9 Back Propagation

[10] Kernel Based Least Mean Square

[11] High Dimensional Space
[12] KM
[13] Support Vector Machines
[14] Kernel Principal Component Analysis
[15] Kernel Independent Component Analysis
[16] LMS
[17] Reproducing Kernel Hilbert Spaces

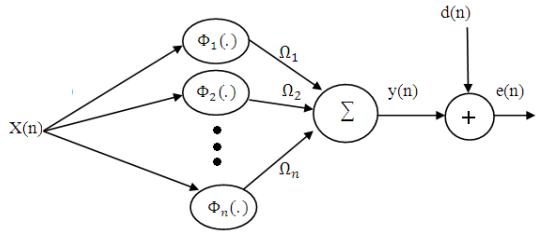

Figure 2: Block diagram of a simple kernel estimation system

### C. Kernel LMS

Estimation and prediction of time-series could be optimized with a new approach, named KLMS [45]. The basic idea is to perform linear LMS algorithm given by equation (3) in the kernel space.

$$\Omega_{(n+1)} = \Omega_{(n)} + 2\mu \times e_{(n)} \times \Phi_{(u_{(n)})} \quad (3)$$

Where $\Omega(n)$ is weight vector in the HDS. The estimated output y(n) will be calculated by equation 4:

$$y_{(n)} = <\Omega_{(n)}, \Phi_{(u_{(n)})}> \quad (4)$$

Figure 2 shows the input vector u(n) being transformed to the infinite feature vector Φ(u(n)), whose components are then linearly combined by the infinite dimensional weight vector. Non-recursive type of Equation (3) can be written as equation (5):

$$\Omega_{(n)} = \Omega_{(0)} + 2\mu \sum_{i=0}^{n-1} e_{(i)} \Phi_{(u_i)} \quad (5)$$

choosing $\Omega(0)=0$:

$$\Omega_{(n)} = 2\mu \sum_{i=0}^{n-1} e_{(i)} \Phi_{(u_{(i)})} \quad (6)$$

From Equations (4) and (6) we derive equation (7):

$$y_{(n)} = <\Omega_{(n)}, \Phi_{(u_{(n)})}> =$$
$$<2\mu \sum_{i=0}^{n-1} e_{(i)}, \Phi_{(u_{(i)})}> = \quad (7)$$
$$2\mu \sum_{i=0}^{n-1} e_{(i)} <\Phi_{(u_{(i)})}, \Phi_{(u_{(n)})}>$$

We can use kernel trick now to calculate y(n) by equation (8):

$$y_{(n)} = \mu \sum_{i=0}^{n-1} e_{(i)} k_{(u_{(i)}, u_{(n)})} \quad (8)$$

Equation (8) is named Kernel LMS. As error of system is reduced by time, one can ignore the e(n) after ξ samples and predict new data with previous error by equation (9):

$$y_{(n)} = \mu \sum_{i=0}^{\xi} e_{(i)} k_{(u_{(i)}, u_{(n)})} \quad (9)$$

This change decreases the complexity of the algorithm. As a result, the system can be trained with fewer data while it is used for prediction of new data.

### D. The Proposed Non-Linear Classifier

As mentioned earlier, KLMS maps input data to actual HDS, followed by linear combining of transformed data as indicated in Fig 2. This obviously is performed based on kernel trick without calculation of weights coefficients.

In the proposed neuron, a nonlinear logistic function is added to KLMS structure for classification purposes. Block diagram of the proposed neural network KLMS is shown in Figure 3 according to which, after transferring input vector $X_{(n)}$ to HDS ($\varphi_{(X(n))}$), a linear combiner mixes all $\varphi_{(X(n))}$ and a nonlinear function f(.) decides which input is to be appointed to which class.

As KLMS is a nonparametric model which learns nonlinear patterns intelligently, we use combination of NN and KLMS to propose a nonparametric classifier. There is no need to initialize parameters of the proposed structure because it adapts itself with new data. The structure of the nonparametric neuron is shown in Figure 4. As indicated, the proposed neuron has two parts: the first part with $\Phi_{i(.)}$ functions map input data $X_{(n)}$ to HDS, while the second part is a perceptron with logistic activation function. With Gaussian $\Phi_{i(.)}$ it would be a radial basis function neural network which is a parametric model. There is no limit in type and number of $\Phi_{i(.)}$ functions. This assumption makes it possible to learn any data pattern. In addition, we assured that any type of data with any distribution would be discriminated in HDS hyper.

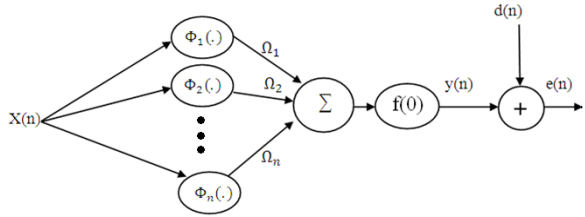

Figure 3: KLMS Neuron structure

Output of the neuron is computed by equation (10) and error is computed by (11):

$$y_{(n)} = f_{(<\Omega_{(n)}, \Phi_{(X_n)}>)} \qquad (10)$$

$$e_{(n)} = d_{(n)} - y_{(n)} \qquad (11)$$

We use KLMS algorithm to train the neuron by equation (12):

$$\begin{aligned}\Omega_{(n+1)} &= \Omega_n + \mu \nabla_{e_n^2} \\ &= \Omega_n + 2\mu \Phi_{X_n} \\ &\times e_n \acute{f}_{(<\Omega_n, \Phi_{(X_n)}>)}\end{aligned} \qquad (12)$$

$E_n$ is defined by equation (13):

$$E_n = e_n \acute{f}_{(<\Omega_n, \Phi_{(X_n)}>)} \qquad (13)$$

$$\Omega_{n+1} = \Omega_n + \mu \nabla_{e_n^2} \qquad (14)$$

$$\Omega_{n+1} = \Omega_n + 2\mu E_n \Phi_{(X_n)} \qquad (15)$$

$$\Omega_n = \Omega_0 + 2\mu \sum_{i=0}^{n-1} E_i \Phi_{(X_i)} \qquad (16)$$

Replacing $\Omega_n$ in $Y_n$, we obtain equation (17):

$$\begin{aligned}y_n &= f_{(<\Omega_n, \Phi_{(X_n)}>)} = \\ &f_{\left(<2\mu \sum_{i=0}^{n-1} E_i \Phi_{(X_i)}, \Phi_{(X_n)}>\right)} = \\ &f_{\left(2\mu \sum_{i=0}^{n-1} E_i <\Phi_{(X_i)}, \Phi_{(X_n)}>\right)}\end{aligned} \qquad (17)$$

Using the kernel, the output will be as equation (18).

$$y_n = f\left(2\mu \sum_{i=0}^{n-1} E_i \times K_{(\Phi_{(X_i)}, \Phi_{(X_n)})}\right) \qquad (18)$$

It turns out that the only information necessary to produce neuron output are $E_i$ coefficients which are determined by equation (13) and are obtained during network training. During the test procedure, for each input data $X_{(n)}$, using $E_i$ and feature vector $X_i$, the classifier output is produced by equation (18).

*E. NNKLMS classifier combiner*

In our new method of classifier fusion, we used a NNKLMS neuron as combiner to fuse outputs of the ensemble of base classifiers. Base classifiers are also neural networks with different parameters and structures. For the combiner to improve performance of base classifiers, it is necessary that the base classifiers are not identical. Structure of the total classifier system is shown in Figure 4. As indicated, BN is the number of base classifiers, CN is the number of classes, and FN is the number of features of input data. Input vector X (1...FN) is an input data for each of the base classifiers. Output of each base classifier is a vector Y(1..C). The value of Y(i) represents the degree of belonging of input X to class i. The outputs of base classifiers would be inputs to the NNKLMS classifier combiner. Therefore, NNKLMS combiner has $BN \times CN$ inputs and CN outputs as a vector NY(1..C) where $NY(i) \in [0..1]$. If $NY(i) = 1$ then the final fusion process decides on input X to belong to class i.

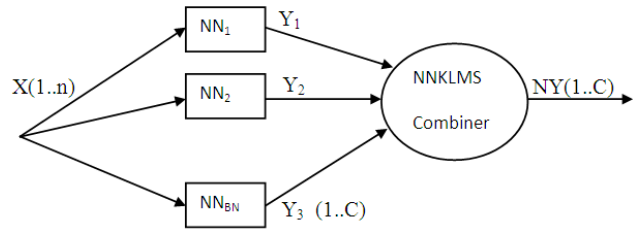

Figure 4: structure of NNKLMS classifier system

## IV. EXPERIMENTAL RESULTS

Experiments were conducted on OCR and seven other benchmark datasets from the UCI Repository (Breast, Wpbc, Iris, Glass, Wine and Heart). As suggested by many classification approaches, some features have been linearly normalized between 0 and 1. Table 1 summarizes some of the features of the datasets.

Table I. Specifications of datasets used

|  | #of samples | #of features | #of classes | window size used | Address |
|---|---|---|---|---|---|
| **OCR** | 1435 | 64 | 10 | 20 | Haghighi@ieee.org |
| **Breast** | 699 | 9 | 2 | 20 | http://archive.ics.uci.edu/ml/machine-learning-databases/breast-cancer-wisconsin/ |
| **Iris** | 150 | 4 | 3 | 20 | http://archive.ics.uci.edu/ml/machine-learning-databases/iris/ |
| **Glass** | 214 | 9 | 6 | 20 | http://archive.ics.uci.edu/ml/machine-learning-databases/glass/ |
| **Wine** | 178 | 13 | 3 | 20 | http://archive.ics.uci.edu/ml/machine-learning-databases/wine/ |
| **Wpbc** | 198 | 32 | 2 | 20 | http://archive.ics.uci.edu/ml/machine-learning-databases/breast-cancer-wisconsin/ |
| **Diabetes** | 768 | 8 | 2 | 20 | http://archive.ics.uci.edu/ml/machine-learning-databases/diabetes/ |

Table II. Output error in NNKLMS vs. other fusion methods.

|  | VT | DS | DTED | DTSD | SM | MAX | PT | MIN | NNKLMS |
|---|---|---|---|---|---|---|---|---|---|
| **OCR** | 3.5 | 3.75 | 3.75 | 3.5 | 3.5 | 4.75 | 4.0 | 5.75 | 2.36 |
| **Breast** | 2 | 2 | 2 | 2 | 2 | 3.5 | 2 | 3.5 | 1.67 |
| **Iris** | 3.33 | 2.22 | 2.22 | 3.33 | 3.33 | 3.33 | 3.33 | 3.33 | 3.33 |
| **Glass** | 20 | 10 | 10 | 10 | 20 | 20 | 20 | 20 | 8.57 |
| **Wine** | 6.87 | 6.87 | 6.87 | 6.87 | 6.87 | 6.25 | 6.87 | 6.87 | 6.38 |
| **Wpbc** | 21.3 | 25.3 | 25.3 | 21.3 | 21.3 | 21.3 | 21.3 | 21.3 | 5.74 |
| **Diabetes** | 26 | 26 | 26 | 26 | 26 | 26 | 26 | 26 | 9.67 |
| **Heart** | 10 | 10 | 10 | 10 | 10 | 10 | 10 | 10 | 8.46 |

One of the most popular benchmarks in image processing applications is OCR. The dataset is a rich database of handwritten Farsi numbers from 0 to 9. Each sample in the database has 64 features indicating the pattern formed by one hand written number based on gray scale of each of the 64 points in a $8 \times 8$ matrix. Figure 5 indicates some of the patterns formed by digits 0 to 9 [46].

Table II compares results of some known methods of classifier fusion with our new NNKLMS method. As seen in the Table, comparison has been carried out with these fusion methods: Voting(VT), Dempster Shafer (DS), Decision template and Euclidean distance(DTED), Decision template and symmetric difference (DTSD), simple mean (SM), maximum (MAX), product (PT), minimum (MIN), average (AV). Specifications of the benchmark datasets have been summarized in Table I.

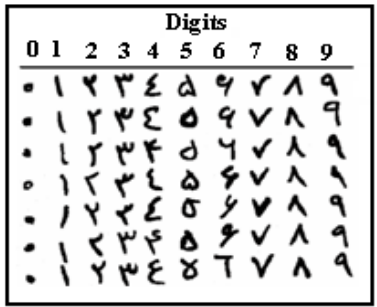

Figure 5: Hand written Farsi numbers

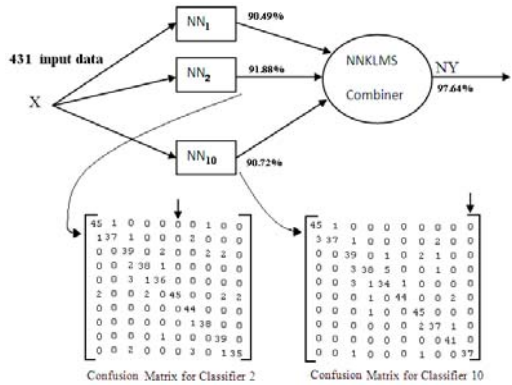

Figure 6: Overall efficiency vs. base classifiers efficiency

Training the combined classifier system has two general steps. First, base classifiers should be trained with training data. It should be noted that the base classifiers used in the combined classifier system should not be identical or have the same behavior on input data because differences in behavior of the classifiers in different regions of the input space is a key element for improving efficiency of the system. For the OCR dataset, 60% of the data were used for training. Therefore, each of the ten base classifiers was trained by these training data. Second, after completion of base classifiers training, it is time to train combiner based on the training data. Since inputs to the combiner are outputs of base classifiers, for each training sample, outputs of all base classifiers are used as a training data for the combiner. As a result, for a system with N base classifier, each of them with output vector of length M, each input data to the combiner would be a N × M featured vector. As a result, for OCR training data, each input sample is fed to each of the 10 base classifiers. Each Output of a base classifier is a vector V of length 10. The value of V[i] shows the degree of belonging of input sample to class i. Outputs of all 10 base classifiers each with length 10 for the same input sample, form an input vector of length 100 as an input training sample for the combiner. For all the training samples, the process continues to train the combiner.

### A. Performance analysis of Base classifiers vs. combiner

Following the fusion process in the classifiers combiner system for one OCR input test data, clearly shows how the combiner system decides based on decisions of ensembles. It is obvious that each of the base classifiers should have different behavior for the input data. To show the difference between them, Figure 6 represents confusion matrix for classifiers 2 and 10 computed for 431 test data. Efficiency of each of the base classifiers and final combiner are also shown in the figure. As confusion matrices show, Efficiency of classifiers 2 and 10 is on maximum value in columns 6 and 10 to distinguish numbers 5 and 9 respectively. Therefore, efficiency of base classifiers 2 and 10 for detecting numbers 5 and 9 is better than others respectively. As a result, efficiency of the combiner improves because each base classifier's efficiency is on maximum value for some of the classes. This different behavior is a key element in more efficient fusion of base classifier outputs by the combiner.

To better investigate efficiency improvement in the NNKLMS combiner with respect to the base classifiers, confusion matrices of classifiers 5 and 6 are compared with confusion matrix of the combiner as shown in Figure 7. In spite of lower efficiency of some of the base classifiers on labeling some of the input data in a number of classes, the combiner shows higher efficiency. Columns 2,3 in confusion matrix of classifier 5 and columns 3,7 in confusion matrix of classifier 6 show low efficiency for labeling data of respective classes while corresponding columns in confusion matrix of the combiner shows higher efficiency. In spite of the fact that some base classifiers may have lower efficiency in some areas of input space; it is expected to have a more efficient classifying system in such fusion process.

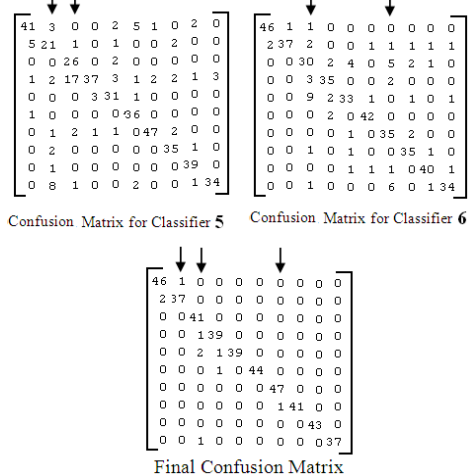

Figure 7: Efficiency improvements in combiner vs. base classifiers

As a general rule, Test procedure is the same as training procedure except that instead of training data, test data is used. One of the problems the test procedure facing with is small number of input data. In such situations, small set of training data causes higher output errors as a result of incomplete training. To overcome the problem, cross validation or leave one out procedures should be used for training and test. In cross validation process, a window of size m < n for input data of size n is defined. The data inside the window are used as test while others outside the window are used as train data. By moving the window on all input data by step size m, then averaging the error in each step, final error rate is computed. Leave one out process is special kind of cross validation where size of the window is limited to 1. Nevertheless, for the data sets where the number of data is less than 500 and output error is more than 10%, cross validation is used. For the data sets with less than 200 samples, leave one out process is used.

## V. CONCLUSION

In this paper, a new method for improving efficiency of combined classifier systems was presented. The proposed method used a special type of neuron named NNKLMS as a combiner. The new neuron used the power of kernel space together with properties of Least means square method to map the problem into a higher dimensional space. As a result, in the new higher dimensional space, more accurate decisions could be made using a two step procedure for training the system. For the datasets with small number of input data, cross validation or leave one out process is used for training and test steps. Results obtained based on benchmark data show that the new combiner is basically more efficient than the other methods. One of the key points for designing such combiner system is that the base classifiers should not be identical because the combiner uses differences in behavior of the base classifiers in different regions of the input space to improve overall efficiency.

About author: Mehdi Salkhordeh Haghighi is studying PhD in computer software in  Computer Department, Ferdowsi University of Mashhad in Iran. He is also a member of teaching staff in Sadjad University in Mashhad, Iran (haghighi@ieee.org).

About author: Abedin Vahedian is associate prof of Computer Department in Ferdowsi University of Mashhad in Iran. He is also a member of teaching staff in the department. (vahedian@um.ac.ir).

About author: Hadi Sadoghi Yazdi is associate prof of Computer Department, in Ferdowsi University of Mashhad in Iran. He is also a member of teaching staff in the department. (sadoghi@sttu.ac.ir).

About author: Hamed Modaghegh is studying PhD in Electrical Engineering in  Electrical Department of  Ferdowsi University of Mashhad in Iran. (modaghegh@yahoo.com).